\pdfoutput=1
\documentclass[11pt]{article}
\usepackage{tikz}
\usetikzlibrary{shapes.geometric, arrows, positioning}
\usepackage[most]{tcolorbox}
\usepackage{caption}
\renewcommand{\thefigure}{\textbf{ \arabic{figure}}}

\tikzstyle{startstop} = [ellipse, minimum width=2cm, minimum height=1cm, text centered, draw=black, fill=red!30]
\tikzstyle{io} = [trapezium, trapezium left angle=70, trapezium right angle=110, minimum width=1cm, minimum height=1cm, text centered, draw=black, fill=blue!30]
\tikzstyle{process} = [rectangle, minimum width=2cm, minimum height=1cm, text centered, draw=black, fill=orange!30]
\tikzstyle{decision} = [diamond, minimum width=2cm, minimum height=1cm, text centered, draw=black, fill=green!30]
\tikzstyle{arrow} = [thick,->,>=stealth]

\usepackage[final]{acl}

\usepackage{times}
\usepackage{latexsym}

\usepackage[T1]{fontenc}
\usepackage[utf8]{inputenc}

\usepackage{microtype}

\usepackage{inconsolata}

\title{Supervisory Prompt Training}

\author{
  Jean Ghislain Billa\footnotemark[1] \\
  MIT \\
  \texttt{jbilla@mit.edu} 
\And
  Min Oh\footnotemark[1] \\
  Microsoft \\
  \texttt{min.oh@microsoft.com} 
\And
  Liang Du \\
  Microsoft \\
  \texttt{liang.du@microsoft.com} \\
  }

\begin{document}
\maketitle
\footnotetext[1]{These authors contributed equally to this work. Corresponding author: min.oh@microsoft.com}
\begin{abstract}
The performance of Large Language Models (LLMs) relies heavily on the quality of prompts, which are often manually engineered and task-specific, making them costly and non-scalable. We propose a novel approach, Supervisory Prompt Training (SPT). SPT automates the generation of highly effective prompts using a dual LLM system. In this system, one LLM, the generator, performs a task while the other, the corrector, provides feedback and generates improved prompts. In contrast to earlier techniques, both the generator and corrector collaboratively and continuously improve their prompts over time. We also introduce the concept of \textit{impact scores} to measure the sentence-level effectiveness of the prompts. Our method was tested on four benchmarks, testing the level of hallucinations in LLMs. Notably, we were able to increase the accuracy of GPT-4 on GSM8K from 65.8\% to 94.1\% (28.3\% increase). SPT advances LLMs by refining prompts to enhance performance and reduce hallucinations, offering an efficient and scalable alternative to traditional model fine-tuning.
\end{abstract}

\section{Introduction}
Recently, Large Language Models (LLMs) have generated a lot of interest due to their unprecedented performance on a diverse range of tasks \citep{xu-etal-2022-gps, yang2023large}. The proficiency of LLMs such as GPT-4 and Llama-2 in generating high-quality text, and their reasoning capabilities make them exceptional agents in various contexts.

However, such performance highly depends on their prompts, and engineering the effective prompt is often a costly process that is not scalable \citep{lester-etal-2021-power,  chen2023instructzero}. Moreover, the long-established prompt is often brittle to even subtle modifications in LLMs such as minor version changes. 

To tackle those limitations, research focuses on automatically generating optimal prompts for different LLMs and tasks. The research primarily concentrates on two areas: continuous optimization and discrete optimization.

Continuous optimization involves building prompts that are vector embeddings. These embeddings are not necessarily word tokens, which can make them incomprehensible to humans and not interpretable \citep{qin-eisner-2021-learning,wen2023hard,li-liang-2021-prefix}. It is also hard to develop continuous prompts when the parameters of an LLM are not inaccessible; a situation that applies to some of the current models \citep{chen2023instructzero,prasad-etal-2023-grips}. On the other hand, discrete optimization aims at producing textual prompts, which are more interpretable \citep{li-liang-2021-prefix}. Because discrete prompts are text-based, LLMs can be used to generate them. For example, AutoHint by \citealp{sun2023autohint} and Automatic Prompt Optimization (APO) by \citealp{pryzant2023automatic} are frameworks with an LLM solving a task, and another LLM providing feedback. In AutoHint, the feedback is continuously added to the prompt of the original LLM, to help it to avoid mistakes. Conversely, APO uses the feedback as "gradients" to generate better prompts for the LLM to accomplish a task. Although these methods have promising results, their main limitation is that there is no guarantee that the LLM providing feedback has an optimal prompt itself. Because the LLM providing feedback is also limited by its own prompt, the feedback may be sub-optimal, which sets an upper bound on the performance of the LLM relying on the feedback.

The current work focuses on this weakness. We introduce Supervisory Prompt Training (SPT), which enhances the prompts of an LLM - referred to as the generator - in performing a task. This is achieved by utilizing another LLM, termed the corrector, to provide feedback. Unlike previous methodologies, the corrector functions as a running mate, progressively refining its own prompts to offer superior feedback over time. 

We also introduce the idea of an \textit{impact score}. For each sentence in the generator's prompt, the impact score quantifies how much adding this sentence improved the accuracy of the generator on the given task. The impact score informs the corrector about the type of sentences that boost the performance of the generator, which allows it to keep producing such sentences.   

Our system is applied to the problem of hallucinations in LLMs. We test the framework on four benchmarks, aimed at gauging the level of hallucinations of an LLM on multiple-choice tasks. We can increase the accuracy of the generator to as much as 27.1\%\, which validates the ability of the system to decrease hallucinations in LLMs. Our contributions are as follows:

\begin{itemize}
    \item We introduce Supervisory Prompt Training, a framework that continuously improves the prompt of LLMs with training examples, 
    \item We produce high-quality interpretable prompts that can decrease the level of hallucinations of multiple LLMs on multiple-choice tasks, 
    \item We showcase that LLM performance can be notably boosted solely through enhanced prompting, which could be an alternative option for model fine-tuning.
\end{itemize}

\section{Related work}
The field of automatic prompt optimization is mainly split between two different research directions: continuous optimization and discrete optimization.
\subsection{Continuous optimization}
Continuous optimization techniques focus on finding prompts using gradients \citep{lester-etal-2021-power,chen2023instructzero}. They produce soft prompts, which are expressive embeddings not limited to word tokens \citep{qin-eisner-2021-learning,li-liang-2021-prefix,wen2023hard}. Soft prompts are highly expressive but lack interpretability from a human standpoint \citep{lester-etal-2021-power,deng-etal-2022-rlprompt}. Additionally, because building soft prompts requires backpropagation, continuous optimization requires access to the parameters of the LLM, which is sometimes impossible because some LLMs are only available through APIs \citep{pryzant2023automatic,chen2023instructzero,prasad-etal-2023-grips,lester-etal-2021-power}. Furthermore, the use of backpropagation in continuous optimization means that the final soft prompt may not be compatible with other models due to differences in embedding dimensions and representation space \citep{wen2023hard}.

\subsection{Discrete optimization}
Discrete optimization aims to improve the performance of an LLM by finding textual prompts. Various prompting techniques have been explored, including zero-shot prompting and few-shot prompting. Zero-shot prompting involves providing instructions to the LLM, while few-shot prompting includes examples of input-output pairs to demonstrate the task to the LLM \citep{NEURIPS2020_1457c0d6}. Chain-of-thought prompting guides the model towards reasoning about its output, resulting in improved performance \citep{wei2023chainofthought}. Few-shot prompting also gave rise to research on choosing the best few-shot examples to maximize LLM performance \citep{deng-etal-2022-rlprompt}. To reduce the need for human intervention, some works use Reinforcement Learning to create optimal prompts. By building a policy and a reward system, researchers can modify specific tokens to create better prompts \citep{deng-etal-2022-rlprompt}. To make prompting more automatic, a considerable part of the research focuses on using other LLMs to find good prompts. \citealp{sun2023autohint} propose AutoHint, a framework that continually adds feedback to the prompt of an LLM, to make it to avoid mistakes in the future. Similarly, \citealp{zhou2023large} uses a black-box LLM to generate instructions for another LLM to complete a task. \citealp{pryzant2023automatic} introduce Automatic Prompt Optimization, which uses feedback from another LLM as "gradients" to generate better prompts. 

One finding from prompt optimization research is that the an LLM's prompt significantly influences its output \citep{lester-etal-2021-power, chen2023instructzero}. When using LLM-based methods to generate feedback or to create a new prompt, the LLM is limited by its own prompt, potentially leading to suboptimal feedback or prompts and subsequently suboptimal task performance. To address this, the current work focuses on a self-improving system, where the LLM improves its prompt over time, leading to better performance.

\section{Proposed method}

\subsection{Initial setup}
The SPT framework comprises two LLM-based agents, namely a generator denoted as \textit{G} and a corrector denoted as \textit{C}. The generator, initialized with a meta-prompt $p_0$, accepting a set of multi-choice questions \textit{D} as parameters can be denoted as $G(p_i, D)$. A meta-prompt can include instructions for an LLM, as well as the LLM persona information (identity, language behavior, and interaction style) \citep{10.1145/3571730}. The generator outputs a list of answer choices evaluated against true answers. We put aside mistakes $m_{p_i}$ that comprise a list of failed questions and the corresponding wrong answers. On the other hand, the corrector, initialized with a meta-prompt $c_0$, accepting the previous (or initial) meta-prompt of the generator $p_i$ and mistakes $m_{p_i}$ can be denoted as $C(c_i, p_i, m_{p_i})$.

Given a multiple-choice task, we randomly split the data into a training dataset $D_{train}$ and a testing dataset $D_{test}$. It’s important to note that we do not expose the test data to the generator and corrector during the prompt training phase. This is done to ensure a fair evaluation of the model’s performance and to avoid overfitting.

The objective of prompt training is to find a meta-prompt \textit{$p_i^*$} at each epoch $i$, that enables the generator to produce choices with a high degree of accuracy for a given task. Simultaneously, we aim to find a meta-prompt \textit{$c_i^*$} for the corrector at each epoch $i$. These $c_i^*$ not only generate meta-prompts that enhance the generator’s accuracy but also fine-tune its own directives to bolster its correction capabilities.

\subsection{Iteratively updating p (SPT-p)}
Our method first prompts the generator to answer all the questions in the training dataset $D_{train}$. We collect the set of questions on which the generator makes mistakes $m_{p_i}$. Using $m_{p_i}$, the corrector is tasked to generate $n$ candidate meta-prompts $p_i^{(1)},p_i^{(2)},...,p_i^{(n)}$ that could help the generator avoid these mistakes in the future. These candidates are then tested as meta-prompts for the generator $G(p_i, D)$ on the questions in the set of mistakes $m_{p_i}$. The candidate that has the best accuracy on this set of questions (so that has been able to make the generator avoid the most mistakes possible) is chosen as $p_i^*$, the best generator meta-prompt for epoch $i$ (\autoref{fig:spt-p}). We then evaluate $p_i^*$ on the testing dataset $D_{test}$.  
 \\
 \\
\begin{tikzpicture}[node distance=1cm, scale=0.7, every node/.style={scale=0.7}]
% Nodes
\node (in1) [io] {\(D_{train}\)};
\node (start) [startstop,right=of in1] {\( p_i \)};
\node (pro1) [process, below=of in1, yshift=-0.5cm] {Generator ($G$)};
\node (dec1) [io, below=of pro1] {Mistakes ($m_{p_i}$)};
\node (pro2) [process, right=of pro1] {Corrector ($C$)};
\node (stop) [startstop, below=of pro2] {\( p_i^* \)};
\node (start2) [startstop, right=of start] { \( c_i \)};

% Arrows
\draw [arrow] (start) -- (pro1);
\draw[arrow]  (start) -- (pro2);
\draw [arrow] (in1) -- (pro1);
\draw [arrow] (pro1) -- (dec1);
\draw [arrow] (dec1) -- (pro2);
\draw [arrow] (pro2) -- (stop);
\draw [arrow] (start2) -- (pro2);
\end{tikzpicture}
\captionof{figure}{\textbf{Flowchart illustrating the iterative process of improving a generator's meta-prompt.} Pink circles represent the meta-prompts in the process, purple trapezes represent data, and orange rectangles represent LLMs.}
\label{fig:spt-p}

\subsection{Iteratively updating p and c (SPT-pc)}
Our premise is that the corrector $C(c_i, p_i, m_{p_i})$ could, over time, generate meta-prompts that keep enhancing the generator's accuracy, but also fine-tune its own directives to improve its correction capabilities. By guiding the corrector to analyze the repeated mistakes the generator makes using $p_{i}$, and subsequently adjusting its own meta-prompt $c_i$, it can direct itself to avoid generating prompts that have the same deficiencies as $p_{i}$. The goal is that the corrector criticizes and improves itself, by giving itself enough information about how to make better $c_i^*$ (\autoref{fig:spt-pc}).  
\\
\\
\begin{tikzpicture}[node distance=1cm, scale=0.7, every node/.style={scale=0.7}]
% Nodes
\node (start) [startstop] {\( c_i \)};
\node (in1) [io, right=of start, xshift=-1cm] {Mistakes ($m_{p_i}$)};
\node (pro1) [process, below=of start] {Corrector ($C$) };
\node (p) [startstop, left= of start] {\( p_i^* \)};
\node (stop) [startstop, below=of pro1] {\( c_i^* \)};
% Arrows
\draw [arrow] (start) -- (pro1);
\draw [arrow] (in1) -- (pro1);
\draw [arrow] (p) -- (pro1);
\draw [arrow] (pro1) -- (stop);
\end{tikzpicture}
\captionof{figure}{\textbf{Flowchart illustrating the iterative process of improving a corrector's meta-prompt.}}
\label{fig:spt-pc}

\subsection{Extended Methods}
Further approaches have been investigated to enhance the performance of both the generator and the corrector. These methods aim to optimize the functionality of the generator and corrector, allowing them to work more effectively and efficiently.

\subsubsection{Iteratively improving p and c with chain-of-thought reasoning (SPT-cot) }
When asking the corrector $C(c_i, p_i, m_{p_i})$ to improve $p_i$, we explicitly ask it to perform step-by-step reasoning over $m_{p_i}$, to understand why every mistake was made. The rest of the procedure remains the same. 

\subsubsection{Impact scores (SPT-imp)}
To keep track of the learnings the corrector gained while improving $p_i$, we introduce the \textit{impact score}. The impact score is a measure of how much adding a sentence to $p_i$ increases the training accuracy of the generator. The impact score is calculated by taking the difference between the accuracies before and after adding the new sentence (\autoref{fig:impact_score}). The impact scores are added at the end of each sentence in $p_i$ with an impact score tag (e.g."You are a useful assistant"; impact score: 0.2") and then, passed into the corrector with meta-prompt $c_i$ and mistakes $m_{p_i}$ when improving $p_i$ so that it has a view of the sentences that increase or decrease the training accuracy. The impact scores are also passed into the corrector when improving $c_i$ so that the corrector has a view of what sentences $c_i^*$ should output so that it creates better $p_i^*$ in the future. 
\\
\\
\begin{tikzpicture}[node distance=1cm, scale=0.5]
% Nodes
\node (start) [startstop, xshift=-2cm,style={scale=0.5}] {$p_0$: '', Accuracy:$0.5$};
\node (cur) [startstop, right=of start,xshift=-0.25cm,style={scale=0.5}] {$p_1$: 'You are an AI assistant',Accuracy:$0.7$};
\node (in1) [io, below=of cur,xshift=-2cm,style={scale=0.5}] {impact scores: \{"You are an AI assistant":$0.2$\}};
% Arrows
\draw [arrow] (start) -- (cur);
\draw [arrow] (start) -- (in1);
\draw [arrow] (cur) -- (in1);
% Label below the diagram
\end{tikzpicture}
\captionof{figure}{\textbf{Flowchart illustrating the iterative process of generating impact scores.}}
\label{fig:impact_score}

\section{Experiments}
In our experiments, we explore the capabilities of SPT using different generators, correctors, and settings. Throughout the experiments, we assess the zero-shot accuracy of the generator model.

\subsection{Problem}
We apply SPT to the problem of hallucinations in LLMs. Hallucinations refer to the phenomena of LLMs generating false predictions with high confidence \citep{sun2023pushing}. This can take the form of outputting non-factual information, or non-sensical information \citep{10.1145/3571730,10.1609/aaai.v37i11.26596}. We focus on the problem of non-factual outputs. We test the ability of SPT to make LLMs output correct answers in different multiple-choice tasks.

\subsection{Models}
We use GPT-3.5-turbo (0314), GPT-4 (0314), and Llama2-70b-chat as generators and  GPT4 (0314) as a corrector. The initial generator meta-prompt $p_0$ is the empty string, and the initial corrector meta-prompt is $c_0$: \textit{"You are an AI expert. You can generate new meta-prompts for another LLM so that this LLM is better at answering questions."}

\begin{table*}
\centering
\begin{tabular}{|l|c|c|c|c|c|c|}
\hline
\textbf{Dataset} & GPT-3.5-turbo & APO & SPT-p & SPT-pc & SPT-cot & SPT-imp\\
\hline
TruthfulQA & 64 & 73.1 & 73.1 & 72.5 & \underline{73.7} & \textbf{79.2}\\
\hline
GSM8K & 55.7 & 54.2 & \underline{57.3} & 56.3 & \textbf{59.8} & 55.9\\ 
\hline
MMLU & \underline{50.5} & \textbf{61.5} & 49.5 & 49.5 & \underline{50.5} & \underline{50.5}\\
\hline
MedQA & \textbf{60.1} & \underline{58.1} & 57.1 & 57.8 & 57.1 & 57.8\\
\hline
\end{tabular}
\caption{Testing accuracies with a GPT-3.5-turbo generator. The best accuracies are in bold, and the second-best accuracies are underlined.}
\label{tab:testing_gpt3}
\end{table*}

\begin{table*}
\centering
\begin{tabular}{|l|c|c|c|c|c|c|}
\hline
\textbf{Dataset} & GPT-4 & APO & SPT-p & SPT-pc & SPT-cot & SPT-imp\\
\hline
TruthfulQA & 81.7 & \textbf{92} & 87.1 & \textbf{92} & 89.6 & 87.1\\
\hline
GSM8K & 65.8 & 68.8 & 89.6 & \textbf{94.1} & \underline{92.9} & 74.5\\ 
\hline
MMLU & 79.7 & 79.6 & . & . & \textbf{80.1} & \underline{79.8}\\
\hline
MedQA & 78.4 & \textbf{79.3} & \underline{78.7} & 77.3 & 77.6 & 78\\
\hline
\end{tabular}
\caption{Testing accuracies with a GPT-4 generator. The best accuracies are in bold, and the second-best accuracies are underlined. "." represents a setting that was unable to be tested.}
\label{tab:testing_gpt4}
\end{table*}

\begin{table*}
\centering
\begin{tabular}{|l|c|c|c|c|c|c|}
\hline
\textbf{Dataset} & Llama2-chat-70b & APO & SPT-p & SPT-pc & SPT-cot & SPT-imp\\
\hline
TruthfulQA & 55.4 & \textbf{68.9} & 65.2 & \underline{67.6} & . & 65.8\\
\hline
GSM8K & 40.3 & \underline{41} & . & . & \textbf{41.3} & 40.9\\ 
\hline
MMLU & \textbf{54.5} & . & . & . & \underline{53} & .\\
\hline
MedQA & \underline{41.6} & . & \textbf{46.1} & . & . & .\\
\hline
\end{tabular}
\caption{Testing accuracies with a Llama2-chat-70b generator. The best accuracies are in bold. "." represents a setting that was unable to be tested.}
\label{tab:accents}
\end{table*}

\subsection{Benchmarks}
Our evaluation benchmarks are TruthfulQA, GSM8K, MMLU, and MedQA. 
\begin{itemize}
    \item TruthfulQA is a benchmark that comprises 817 questions that span 38 categories, including health, law, finance, and politics. The benchmark tests whether an LLM outputs truthful information given a question \citep{lin2022truthfulqa}. For this experiment, TruthfulQA was split into 653 training examples and 164 testing examples.
    \item GMS8K is a grade school math word problem benchmark with 7,473 training samples and 1,319 test samples \citep{cobbe2021training}.
    \item MMLU tests the extent of the world knowledge of an LLM over 57 tasks \citep{hendrycks2021measuring}. For this experiment, we have 1816 training examples, and 14042 test examples. 
    \item MedQA focuses on medical problems. We test our system on MedQA-US 4 options. It contains 12,723 questions in English, that we use in the current experiment after a train-test split of 11450 training examples, and 1273 testing examples.
\end{itemize}
The benchmarks were adapted to be able to measure the level of hallucinations in the generator models. They were all used in a multiple-choice setting with 3 answer choices, except for MMLU and MedQA which have 4 answer choices. Only one of the answer choices is the correct answer. The model choosing an incorrect answer is considered a hallucination.

\subsection{Comparison}
As a baseline, we tested base generative models without prompt engineering (i.e. GPT-3.5-turbo (0314), GPT-4 (0314), and Llama2-70b-chat). Also, we used Automatic Prompt Optimization by \citealp{pryzant2023automatic}. To the best of our knowledge, this work is state-of-the-art (SOTA) in terms of LLM-based automatic prompting, and we show that improving the corrector can have better results than just improving a generator.

\section{Results and Discussion}

\subsection{GSM8K}
On the GSM8K dataset, the increase in accuracy of all the generator models seems to be dependent upon the mathematical capabilities of the model. The most powerful model we use as a generator (GTP-4) has an increase in testing accuracy of \textbf{$28.3\%$}, better than the SOTA method. GPT-3.5-turbo has an increase in accuracy of \textbf{$4.1\%$}, also better than the SOTA method. The Llama2-chat-70b generator has the smallest increase in accuracy, \textbf{$1\%$}. This trend in improvements may mean that the efficacy of SPT may be contingent upon the inherent reasoning capabilities of the generator model. Although the corrector produces a good prompt, the performance still depends on the intrinsic abilities of the LLM.

Analyzing the prompts also provides insights into the accuracy boost. The strength of the generator prompt produced (\autoref{fig:gpt4GSM8K}) lies in specificity. Because the corrector $C(c_i, p_i, m_{p_i})$ was able to analyze the past mistakes of the generator, it gave specific feedback, which helped the generator do better on the task. The specific guidelines are about the model being careful and correct when answering questions, while particularly considering special aspects of each question. 

The specificity of the generator prompt emphasizes how essential the feedback from the corrector is, which justifies the need to make the corrector create the best feedback possible. A sample of a corrector meta-prompt (\autoref{fig:correctorGSM8K}) shows how the corrector was encouraged to pay attention to mistakes and to keep refining its ability to create specific meta-prompts. 

\subsection{TruthfulQA}
The results on TruthfulQA follow a different trend than the ones on the GSM8K. Indeed, the testing accuracy increases by \textbf{$10.3\%$} on GPT4, \textbf{$15.2\%$} on GPT-3.5-turbo, and \textbf{$11.8\%$} on Llama2-70b-chat. Analyzing the prompt of the GPT-3.5 generator (\autoref{fig:gpt35TruthfulQA}) sheds some light on the accuracy boost. The prompt produced is specific and provides clear guidelines for the generator to improve. Those guidelines push the model towards focusing on latest information while avoiding misconceptions. Given the range of multiple-choice questions in TruthfulQA, these guidelines do make sense. Additionally, guiding the model to not answer as if it had personal experiences refers to multiple questions in TruthfulQA where the correct answer is "I have no comment". This guideline then enforces the fact that the model should not hallucinate and try to answer questions related to personal experiences. Generating such specific feedback was driven by the ever-improving corrector (\autoref{fig:correctorTruthful}). The corrector's meta-prompt considers the complexities of the questions and pushes the corrector to identify them and include them in the meta-prompts it generates.

\subsection{MMLU}
On the MMLU dataset, the best result is an increase in testing accuracy of \textbf{$0.4\%$} using a GPT-4 generator. We hypothesize that the results have not been satisfying across the board because MMLU has the smallest ratio of training versus testing data (ratio of 0.13; 1,816 training examples vs 14,042 testing examples). We had decided to train on a smaller amount of data because of time and resource constraints. Training on a smaller amount of data points was also a way to assess how much training data is needed for SPT to work properly. We hypothesize that the small information contained in the training data was not enough for the corrector to be able to build generalizable $p^*$, which led to a poorer performance on the testing dataset. However, the generator and corrector prompts still show signs of learning and improvement. Through the generator's meta-prompt (\autoref{fig:gpt4genMMLU}), we see the diverse topics of the MMLU dataset. Once again, the corrector produced the generator prompt by focusing on specificity. The corrector prompt can be further analyzed in \autoref{fig:gpt4MMLU}.

\subsection{MedQA}
The best result we obtained on MedQA was a \textbf{$5\%$} increase in accuracy using a Llama2-70b-chat generator. The Llama2-70b-chat generator's meta-prompt can be further analyzed in \autoref{fig:llama-medqa}. We again see one of the particularities of SPT: specific prompts that guide the generator in the right direction, improving its accuracy. For both GPT-3.5-turbo and GPT-4 though, SPT does not perform as well as the baseline.  

\section{Limitations}  
\label{sec:limitations}  
  
While SPT offers significant advancements in the field of LLMs, it is not without limitations and potential risks that should be acknowledged and addressed.
The prompts generated through SPT are highly effective within the specific context of the training data. However, their generalizability to unseen data is not guaranteed. The prompts may overfit the characteristics of the training set, leading to reduced performance in different settings.
The success of SPT is contingent on the availability and quality of training data. Inaccuracies or biases in the training data can be amplified through the prompt training process, potentially leading to suboptimal or skewed model outputs.
The iterative nature of SPT may result in lengthy and complex prompts that are challenging to interpret. This can obscure the decision-making process of the LLMs and complicate the task of troubleshooting or refining the prompts.  
SPT is resource-intensive, requiring significant computational power and time to iterate over multiple prompt candidates. This may limit its accessibility and scalability, especially for users with constrained resources.
Despite efforts to reduce hallucinations—instances where LLMs generate false predictions with high confidence—there is an inherent risk that SPT may not fully eliminate such occurrences. Hallucinations can have serious implications, particularly in high-stakes domains such as healthcare or legal advice.
As a side note, Prompt engineering may encode human biases into LLMs, so ethical considerations are essential to prevent perpetuating or exacerbating societal biases.  
In light of these limitations, ongoing research and development are crucial to enhance the robustness, ethical soundness, and real-world applicability of prompt-based training methods for LLMs.

\section{Conclusion}  
We present SPT, a novel method for improving LLMs. SPT improves performance on LLM hallucinations benchmarks, but its effectiveness relies on training data and LLM capabilities. SPT may produce long prompts and demands significant computational resources. Future work could explore the impact of different corrector models, prompt length limits, and broader applicability across tasks.  

\bibliography{anthology,custom}

\section*{Appendix}
\appendix
\renewcommand{\thefigure}{A.\arabic{figure}}
\setcounter{figure}{0}

\begin{figure*}
    \centering
    \includegraphics[width=0.4\textwidth]{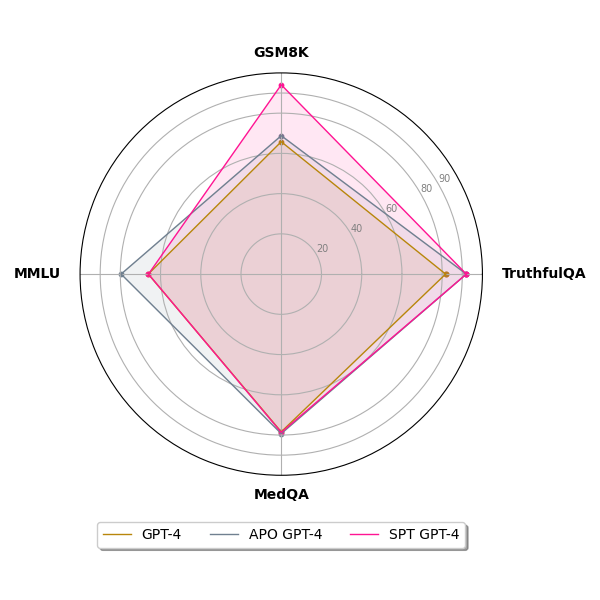}
    \includegraphics[width=0.4\textwidth]{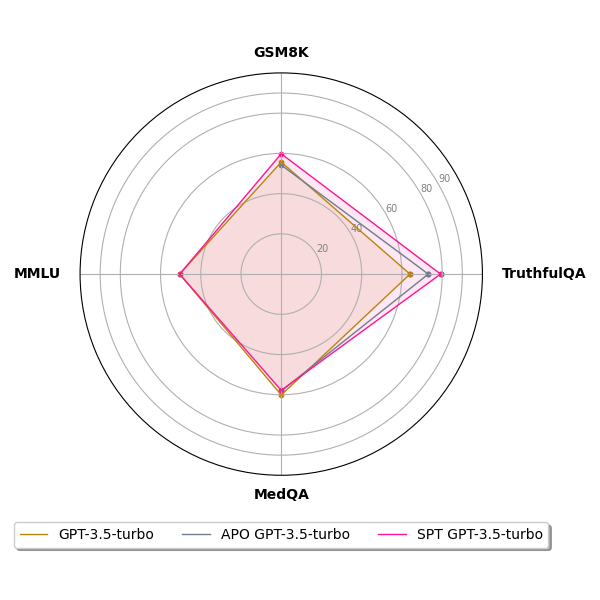}
    \includegraphics[width=0.5\textwidth]{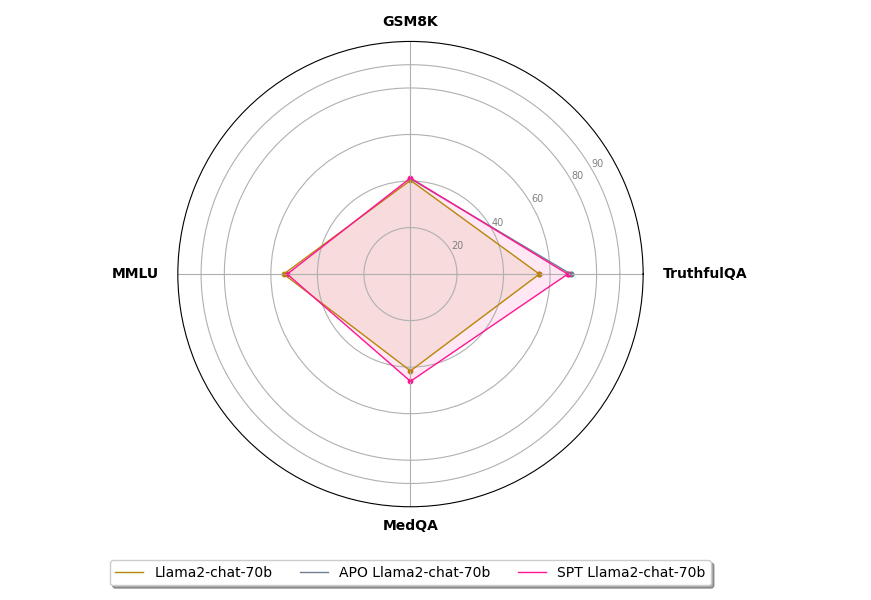}
    \caption{Comparison of SPT to the generator's baseline and APO. SPT does consistently better than the baselines on TruthfulQA and GSM8K.}
    \label{fig:three_figures}
\end{figure*}

\section{Additional meta-prompts}

\begin{figure*}
\begin{tcolorbox}[colback=pink!5!white, colframe=pink!75!black, title=Sample GPT4 Generator's meta-prompt for GSM8K, width=\linewidth, breakable]
...provide \textbf{accurate and precise answers}...Carefully analyze each question and its context... When answering questions about calculations, ratios, percentages, and other numerical topics, ensure that your answers are \textbf{mathematically accurate and logically consistent. Double-check your calculations and reasoning to avoid common mistakes}...be mindful of the relationships between different elements in the questions, such as proportions, timeframes, and sequences...When dealing with multi-step problems, ensure that you \textbf{follow each step correctly} and consider the implications of previous steps on the current step... In cases where the question involves real-life scenarios or practical applications, make sure to consider any constraints or limitations that may be present, and adjust your calculations and reasoning accordingly to \textbf{provide the most accurate and realistic answer possible}...
\end{tcolorbox}
\caption{\textbf{GPT4 Generator's meta-prompt for GSM8K}}
\label{fig:gpt4GSM8K}
\end{figure*}

\begin{figure*}
\begin{tcolorbox}[colback=pink!5!white,colframe=pink!75!black,title=Sample GPT4 Corrector meta-prompt for GSM8K,width=\linewidth, breakable]
...As an \textbf{AI expert}, focus on generating better meta-prompts for other LLMs by analyzing the mistakes made in previous meta-prompts, \textbf{identifying the root causes of these errors}, and incorporating this knowledge into the creation of improved meta-prompts. \textbf{Continuously refine your ability to create better meta-prompts} by learning from the successes and failures of previous LLMs, and apply this knowledge to enhance the problem-solving capabilities of future LLMs
\end{tcolorbox}
\caption{\textbf{GPT4 Corrector's meta-prompt for GSM8K}}
\label{fig:correctorGSM8K}
\end{figure*}

\begin{figure*}
\begin{tcolorbox}[colback=pink!5!white,colframe=pink!75!black,title=Sample GPT-3.5 turbo Generator's meta-prompt for TruthfulQA,width=\linewidth, breakable]
...prioritize delivering \textbf{accurate, well-researched, and up-to-date information}, while demonstrating cultural sensitivity and awareness of the latest developments. Focus on addressing users' needs with clarity and precision, avoiding biases, assumptions, and stereotypes...\textbf{Pay close attention to specific details and keywords} in each question to ensure relevance and accuracy...In cases involving legal or ethical issues, base your information on existing laws and regulations, avoiding judgments or endorsements of potentially harmful actions...For questions involving personal experiences, emotions, or future events, acknowledge the uncertainty or \textbf{refrain from answering as if you possess personal experiences or emotions}.
\end{tcolorbox}
\caption{\textbf{GPT-3.5 turbo Generator's meta-prompt}}
\label{fig:gpt35TruthfulQA}
\end{figure*}

\begin{figure*}
\begin{tcolorbox}[colback=pink!5!white,colframe=pink!75!black,title=Sample GPT4 Corrector's meta-prompt for TruthfulQA,width=\linewidth, breakable]
As an AI expert, \textbf{generate superior meta-prompts for other LLMs}, ensuring they provide precise, well-researched, and current information in response to questions. \textbf{Emphasize critical thinking, skepticism, and a profound understanding of context and nuances. Continuously improve your ability to create enhanced meta-prompts} for other LLMs, addressing any shortcomings and boosting their performance in answering questions. Pay particular attention to the mistakes made in previous responses and \textbf{ensure that the new meta-prompts guide LLMs to avoid repeating those errors}, providing accurate and reliable information. Focus on refining your understanding of question phrasing, context, and potential pitfalls in interpretation to generate more accurate and reliable meta-prompts.
\end{tcolorbox}
\caption{\textbf{GPT4 Corrector's meta-prompt for TruthfulQA}}
\label{fig:correctorTruthful}
\end{figure*}

\begin{figure*}
\begin{tcolorbox}[colback=pink!5!white,colframe=pink!75!black,title=Sample GPT4 Generator's meta-prompt for MMLU,width=\linewidth, breakable]
...\textbf{prioritize understanding the context, details, and any mathematical, logical, or ethical operations} required by the question. Ensure that your \textbf{response is accurate, relevant}... Be mindful of \textbf{potential pitfalls or common misconceptions}...Evaluate each element and its relationship to the others in questions involving comparisons, sequences, or sets...Strive for precision and clarity in your answers, and be prepared to adapt your approach based on the specific requirements of each question. Carefully consider the logical structure and implications of each statement when evaluating statements for truth or falsehood...
\end{tcolorbox}
\caption{\textbf{GPT4 Generator's meta-prompt for MMLU}}
\label{fig:gpt4genMMLU}
\end{figure*}

\begin{figure*}
\begin{tcolorbox}[colback=pink!5!white,colframe=pink!75!black,title=Sample GPT4 Corrector's meta-prompt for MMLU,width=\linewidth, breakable]
...\textbf{prioritize generating improved meta-prompts for other LLMs by thoroughly understanding the context, details, and any mathematical, logical, or ethical operations required by the questions}. \textbf{Focus on addressing the specific mistakes and shortcomings of previous meta-prompts...and use this information to enhance the accuracy, relevance, and comprehensiveness of the new meta-prompts}...Ensure that the new meta-prompts guide the LLMs to be mindful of potential pitfalls or common misconceptions that may lead to incorrect answers, and take care to avoid them in their responses...emphasize the importance of cross-referencing with relevant knowledge or principles, and being prepared to adapt the approach based on the specific requirements of each question...
\end{tcolorbox}
\caption{\textbf{GPT4 Corrector's meta-prompt for MMLU}}
\label{fig:gpt4MMLU}
\end{figure*}

\begin{figure*}
\begin{tcolorbox}[colback=pink!5!white,colframe=pink!75!black,title=Sample Llama2-chat-70b's meta-prompt for MedQA,width=\linewidth, breakable]
...your primary task is to provide \textbf{accurate and comprehensive responses to a wide range of medical questions}, involving clinical scenarios, patient histories, laboratory results, and medical imaging. Pay \textbf{close attention to specific details in each question}, such as symptoms, medical history, test results, patient's age, and the timeline of the symptoms...consider the genetic, environmental, and lifestyle factors that may influence the patient's condition. Be aware of the potential for rare diseases or conditions in patients with unusual or unexplained symptoms. When considering treatment options, take into account the potential side effects and interactions of medications, as well as the patient's personal and cultural preferences. In addition, when dealing with questions about medical procedures or interventions, consider the most likely outcomes and potential complications based on the patient's specific condition and overall health status. When dealing with questions about psychiatric disorders, consider the age of onset, duration, and specific criteria for the diagnosis. For questions about metabolic disorders, consider the specific metabolic pathway affected and the corresponding clinical manifestations. When dealing with questions about patient management, consider the most appropriate next step in management based on the patient's specific condition and overall health status...When dealing with questions about medical emergencies, consider the most appropriate immediate response and the necessary steps for preserving the patient's health and safety. When dealing with questions about pregnancy and prenatal care, consider the potential impact of the mother's health and lifestyle on the fetus, as well as the appropriate monitoring and interventions based on the stage of pregnancy and the mother's health status. When dealing with questions about endocrine disorders, consider the specific hormones affected and the corresponding clinical manifestations and potential complications. When dealing with questions about surgical procedures, consider the most appropriate method for preserving the integrity of the tissue and the patient's overall health status. When dealing with questions about drug mechanisms, ensure you understand the pharmacodynamics and pharmacokinetics of the drug in question. When dealing with questions about genetic conditions, consider the inheritance patterns and potential complications of the condition. When dealing with questions about acid-base disorders, ensure you understand the underlying pathophysiology and compensatory mechanisms.
\end{tcolorbox}
\caption{\textbf{ Llama2-chat-70b's meta-prompt for MedQA}}
\label{fig:llama-medqa}
\end{figure*}

\section{Implementation details}
We present the full prompt of the generator when asked a question on a dataset. To ensure that the generator only outputs the number of the correct answer in a multiple choice setting, we use a guidance program detailed in \autoref{fig:generator_prompt}.  We also describe the full prompts when querying the corrector to modify the generator prompt $p_i$.

\begin{figure*}
\begin{tcolorbox}[enhanced, breakable, colback=pink!5!white,colframe=pink!75!black, width=\textwidth, before=\hfill, after=\hfill,title= Generator's full prompt]
\begin{verbatim}

'''{{#system~}}
    {{p_i}}
    {{~/system}}
                                            
    {{#user~}}
    {{question}}
    The correct answer is:
    {{~/user}}
                                            
    {{#assistant~}}
    {{gen 'answer' temperature=0}}
    {{~/assistant}}
    
    {{#user~}}
    Therefore, the number of the  correct answer is:
    {{~/user}}

    {{#assistant~}}
    {{select 'answer_choice' options=numeric_choices}}
    {{~/assistant}}'''
\end{verbatim}

\end{tcolorbox}
\caption{\textbf{Full generator guidance program}. numeric choices is an array of are numbers from 1-4 or 1-3 depending on the number of answer choices.}
    \label{fig:generator_prompt}
\end{figure*}

\begin{figure*}
\begin{tcolorbox}[enhanced, breakable, colback=pink!5!white,colframe=pink!75!black, width=\textwidth, before=\hfill, after=\hfill,title= Corrector's full prompt - Iteratively improving p (SPT-p)]
\begin{verbatim}
[
    {
        "role": "system",
        "content": {{c_i}},
    },
    {
        "role": "user",
        "content": "Here is a list of questions, answers generated by an LLM 
                    and the correct answers. Next, you have the meta-prompt 
                    of the LLM. The LLM made mistakes on these questions 
                    because of this meta-prompt. Generate an excellent 
                    meta-prompt for the LLM so it can find the correct answer 
                    for all the questions. You must understand every single 
                    question, with every single wrong answer given by the 
                    other LLM, and understand why the other LLM answered with 
                    a wrong answer. You must pay attention to all the 
                    questions' topics and you must ensure the new meta-prompt 
                    clearly explains how the LLM should go about answering all 
                    the questions about those topics correctly. You must also 
                    keep the important ideas in the current meta-prompt intact. 
                    Only output the new meta prompt preceded by 'New prompt: '.
                    List of questions: {{List of questions}}; Original LLM 
                    meta prompt: {{$p_i$}}."
    }
]
\end{verbatim}
\end{tcolorbox}
\caption{\textbf{Full corrector prompt - Iteratively improving p}}
\end{figure*}

\begin{figure*}
\begin{tcolorbox}[enhanced, breakable, colback=pink!5!white,colframe=pink!75!black, boxsep=5pt, arc=5pt, width=\textwidth, before=\hfill, after=\hfill,title= Corrector's full prompt - Iteratively improving p and c (SPT-pc)]
\begin{verbatim}
[
    {
        "role": "system",
        "content": {{c_i}},
    },
    {
        "role": "user",
        "content": "You generate better meta-prompts for other LLMs, and these new 
                    meta-prompts solve all the mistakes of the LLM. You accomplished 
                    it for other LLMs using your meta-prompt c_0: {{c_i}}. However, 
                    the initial meta-prompt of an LLM p_0: "{{p_i}}" and the new 
                    meta-prompt p*: "{p_i*}" that you generated made mistakes on the 
                    same questions: {{m_pi}}. Generate a new meta-prompt for yourself 
                    that is better than c_0, that must create better meta-prompts than 
                    p* in the future. You must ensure that the new meta-prompt strongly 
                    emphasizes your ability to create better meta-prompts for other 
                    LLMs, taking into account the aforementioned mistakes. Only 
                    output the new meta prompt preceded by 'New prompt:"
    }
]
\end{verbatim}
\end{tcolorbox}
\caption{\textbf{Full corrector prompt - Iteratively improving p and c}}
\end{figure*}

\begin{figure*}
\begin{tcolorbox}[enhanced, breakable, colback=pink!5!white,colframe=pink!75!black, boxsep=5pt, arc=5pt, width=\textwidth, before=\hfill, after=\hfill,title= Corrector's full prompt - Iteratively improving p with chain-of-thought reasoning (SPT-cot)]
\begin{verbatim}
[
    {
        "role": "system",
        "content": {{c_i}},
    },
    {
        "role": "user",
        "content": "Here is a list of questions, answers generated by an LLM and the
                    correct answers. Next, you have the meta prompt of the LLM. The 
                    LLM made mistakes on these questions because of this meta-
                    prompt. First, do a step-by-step reasoning on all the problems 
                    with the current prompt that made the LLM fail at finding the 
                    right answers. You must understand every single question, with 
                    every single wrong answer given by the other LLM, and understand 
                    why the other LLM answered with a wrong answer. Then, generate 
                    an excellent meta prompt that resolves all those problems. You 
                    must pay attention to all the questions' topics. Output the new 
                    meta prompt preceded by 'New prompt: '. List of questions: 
                    {{List of questions}}; Original LLM meta-prompt: {{$p_i$}}"
    }
]
\end{verbatim}

\end{tcolorbox}
\caption{\textbf{Full corrector prompt - Iteratively improving p with chain-of-thought reasoning}}
\end{figure*}

\begin{figure*}
\begin{tcolorbox}[enhanced, breakable, colback=pink!5!white,colframe=pink!75!black, boxsep=5pt, arc=5pt, width=\textwidth, before=\hfill, after=\hfill,title= Corrector's full prompt - Iteratively improving p with impact score (SPT-imp)]
\begin{verbatim}
[
    {
        "role": "system",
        "content": {{c_i}},
    },
    {
        "role": "user",
        "content": "Here is a list of questions, answers generated by an LLM 
                    and the correct answers. Next, you have the meta-prompt 
                    of the LLM. The LLM made mistakes on these questions 
                    because of this meta-prompt. Generate an excellent 
                    meta-prompt for the LLM so it can find the correct answer 
                    for all the questions. You must understand every single 
                    question, with every single wrong answer given by the 
                    other LLM, and understand why the other LLM answered with 
                    a wrong answer. You must pay attention to all the 
                    questions' topics and you must ensure the new meta-prompt 
                    clearly explains how the LLM should go about answering all 
                    the questions about those topics correctly. You must also 
                    keep the important ideas in the current meta-prompt intact. 
                    Only output the new meta prompt preceded by 'New prompt: '.
                    List of questions: {{List of questions}}; Original LLM 
                    meta prompt: {{$p_i$\}}.
                    Here is a history of sentences and how they impacted the 
                    correctness of the LLM out of 1. You must use this information 
                    to create a better prompt for the LLM.: {{impact scores}}"
    }
]
\end{verbatim}

\end{tcolorbox}
\caption{\textbf{Full corrector prompt - Iteratively improving p with impact score}}
\end{figure*}

\begin{figure*}
\begin{tcolorbox}[enhanced, breakable, colback=pink!5!white,colframe=pink!75!black, boxsep=5pt, arc=5pt, width=\textwidth, before=\hfill, after=\hfill,title= Corrector's full prompt - Iteratively improving c with impact scores (SPT-imp)]
\begin{verbatim}

[
    {
        "role": "system",
        "content": {{c_i}},
    },
    {
        "role": "user",
        "content": "You generate better meta-prompts for other LLMs, and these new 
                    meta-prompts solve all the mistakes of the LLM. You accomplished 
                    it for other LLMs using your meta-prompt c_0: {{c_i}}. Here are 
                    the impact scores of each sentence in another LLM's meta-prompt 
                    out of 1: {{impact scores}} Generate a new meta-prompt for 
                    yourself that is better than c_0 and must create sentences with 
                    higher impact scores than the ones mentioned above. Your new 
                    meta-prompt should allow you to generate meta-prompts that have 
                    sentences that have a high impact score. Only output the new 
                    meta-prompt preceded by 'New prompt:"
    }
]
\end{verbatim}
\end{tcolorbox}
\caption{\textbf{Full corrector prompt - Iteratively improving c with impact score}}
\end{figure*}

\end{document}